# On-the-Fly SfM: What you capture is What you get

Zongqian Zhan, Rui Xia, Yifei Yu, Yibo Xu, Xin Wang

*Abstract* — Over the last decades, ample achievements have been made on Structure from motion (SfM). However, the vast majority of them basically work in an offline manner, i.e., images are firstly captured and then fed together into a SfM pipeline for obtaining poses and sparse point cloud. In this work, on the contrary, we present an on-the-fly SfM: running online SfM while image capturing, the newly taken On-the-Fly image is online estimated with the corresponding pose and points, i.e., *what you capture is what you get*. More specifically, our approach firstly employs a vocabulary tree that is unsupervised trained using learning-based global features for fast image retrieval of newly fly-in image. Then, a robust feature matching mechanism with least squares (LSM) is presented to improve image registration performance. Finally, via investigating the influence of newly fly-in image's connected neighboring images, an efficient hierarchical weighted local bundle adjustment (BA) is used for optimization. Extensive experimental results demonstrate that our on-the-fly SfM can meet the goal of robustly registering the images while capturing in an online way.

## I. INTRODUCTION

Structure from Motion (SfM) has been a pivotal topic in the field of computer vision, robotics, photogrammetry, which are widely applied in augmented reality [1], autonomous driving [2-4], and 3D reconstruction [5]. Heretofore, many impressive SfM approaches have been extensively studied, mainly including Incremental SfM [5-9], Hierarchical SfM [10-13] and Global SfM [14-20], depending on the procedure of how images are registered. However, these SfM methods predominantly operate in an offline manner, i.e., images are firstly captured, feature extracting\matching and epipolar geometry validation are then performed using all images, one specific SfM method is selected to estimate poses of all images and the corresponding sparse point cloud. This conventional offline SfM typically limits the possibility for online measurement, rapid quality evaluation, etc.

In response to real-time performance, there exists another related hot research topic of VSLAM (Visual Simultaneous Localization and Mapping) worth referring to, it can deal with video data in real time. Given sequential frames, VSLAM can compute real-time trajectory of cameras and 3D object points. Generally, with various embedded sensors, VSLAM can be mainly categorized into mono-VSLAM, stereo-VSLAM and Inertial-VSLAM [21-25], they all contain several common modules: *tracking*, inputting frames and outputting the corresponding pose; *local mapping*, generating 3D points and optimizing local maps; *loop closure*, detecting loop and refining loop correction. The inherent assumption of VSLAM requires that the input frames must be spatiotemporally continuous [21], which means two adjacent frames must be contiguous in time and space or auxiliary information from GPS/IMU [23,24] is available, this consequentially hinders the way that the data can be collected.

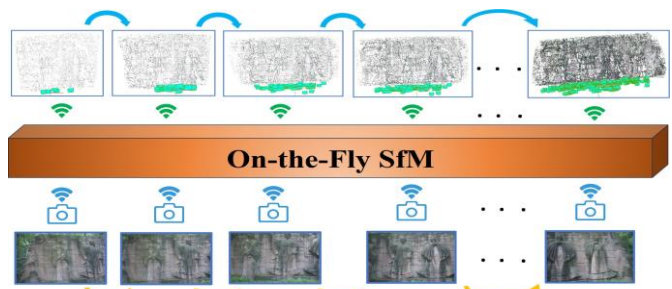

Figure 1. The proposed on-the-fly SfM.

In this paper, as Fig. 1 exemplifies, we present a novel on-the-fly SfM: running online SfM while image capturing. Similar to conventional SfM, on-the-fly SfM yields image poses and 3D sparse points, but we do this while the image capture. More specifically, the current image's pose and corresponding 3D points can be estimated before next image is captured and on-the-fly to be processed, i.e., *what you capture is what you get*. Also, analogous to VSLAM that can ensure real-time performance, on-the-fly SfM is further designed to be able to deal with images captured in an arbitrary way, whereby the spatiotemporal continuity is not necessary any more. The proposed SfM is mainly composed of three steps: online image collecting module, fast image matching and efficient geometric processing. The first one is first established with a camera and a Wifi transmitter, which immediately send the captured image for processing via Wifi signal transmission. The second step is to efficiently and robustly generate the matching results between the already registered images and the new fly-in image, in which fast image retrieval is the most important component for real-time performance. The last step is to estimate camera pose and 3D points robustly and fast, besides the canonical image registration and triangulation, an efficient hierarchical weighted local bundle adjustment is adopted. For each new fly-in image, we just iterate these three steps. More details can be found in Section III-A.

To approach the goal of *what you capture is what you get*, along with the presented on-the-fly SfM using a new online working mode, we also make three technical contributions:

- Fast image retrieval based on learning-based global feature and vocabulary tree. In this work, we extract the global feature using the pre-trained model [26] and unsupervised train vocabulary tree. For each new fly-in image, the global feature is computed and traversed along the vocabulary tree for fast image retrieval.

- Refinement of correspondences using Least Squares Matching [27]. Based on the original matching

*This work was jointly supported by the National Science Foundation of China (No. 61871295, 42301507) and Natural Science Foundation of Hubei Province, China (No. 2022CFB727) and ISPRS Initiatives 2023.

All the authors are with the School of Geodesy and Geomatics, Wuhan University, China P.R., {xiarui, ybxusgg, yfyu2020}@whu.edu.cn, {zqzhan, xwang}@sgg.whu.edu.cn (corresponding author: Rui Xia, Xin Wang).

mechanism (e.g., SIFT [28]), considering the geometric and photometric consistency around the local windows of matched points, a least squares system is applied to refine the 2D position of correspondences according to the grey values within the relevant local windows on two images.

- Hierarchical weighted local BA for efficient optimization of poses and 3D points. For each new fly-in image, only its neighboring connected images (already registered) are enrolled in BA. In addition, based on our image retrieval result, the influence of various connected images on the newly captured image is implied by hierarchical weights, which are employed as priors for improving BA.

## II. RELATED WORKS

In this section, two related topics (SfM and SLAM) are briefly reviewed, which mainly includes some popular works. In addition, some state-of-the-art studies regarding image retrieval and efficient bundle adjustment are introduced.

### A. SfM & VSLAM

So far, there are a lot of open public SfM packages, e.g., VisualSFM [29], OpenMVG[30], Theia[31], Colmap [5], etc. However, all these packages basically concentrate on offline processing mode. For example, Colmap, one of the most widely-used packages, furnishes an end-to-end 3D reconstruction pipeline for large-scale unordered images and it unfolds via a structured pipeline that is mainly comprised of three key stages: image matching, pose estimation and sparse reconstruction, dense reconstruction. To achieve the goal of real-time SfM, inspired by monocular VSLAM, Song et al. [32] presented a monocular SfM that concentrated on eliminating scale drift using the information of ground plane. They yielded comparable performance to stereo setting on long-time sequences. Zhao et al. [33] proposed a so-called real-time SfM (RTSfM), in which feature matching was improved by a hierarchical feature matching strategy based on BoW (Bag-of-Word) [34] and multi-view homography, and a graph-based optimization was employed for efficiency. However, both the reviewed online SfM methods still rely on the spatiotemporal continuity between images or require GPS.

Furthermore, there already exist quite a few mature VSLAM methods that are capable of real-time performance in specific scenarios or tasks. For example, the very popular ORB-SLAM series [21, 24, 25] was continuously published and support a wide range of camera models and sensors, allowing them to achieve high-precision localization while capturing frames, and the VINS-Fusion [35] combined input of images and GPS/IMU which are widely adopted in autonomous driving. However, the robustness of these VSLAM methods is limited in certain scenarios, such as weak textures and motion blur. Yue et al. [27] integrated least squares into the feature matching of ORB-SLAM2 and provided more precise observations. Please note that while there are ample VSLAM methods worth reviewing, this review section only lists a few popular and relevant works.

Contrary to conventional SfM, our proposed on-the-fly SfM is deployed with real-time online processing while image capturing in an arbitrary way. Comparing to VSLAM, the major advantage of our method is that the requirement for input images' spatiotemporal continuity is not necessary any more, nor is the independence of GPS/IMU.

### B. Image retrieval

Image retrieval technique has been widely deployed in SfM and VSLAM for accelerating feature matching and loop closure detection. One typical idea is to build an efficient indexing structure using local features (e.g., SIFT, ORB), in which the BoW is one of the most representative methods to fast identify similar image pairs and loop closure, such as [36]. Similar to BoW, Havlena and Schindler [37] trained a two-layer vocabulary tree for speeding up image matching. Wang et al. [38] introduced random KD-forest consisted of several independent KD-trees, and matchable image pairs can be efficiently determined via traversing on the KD-forest.

In the last few years, learning-based methods have greatly improved image retrieval regarding both time efficiency and precision. Arandjelović et al. [39] proposed a trainable pooling layer via a soft assignment for VLAD, which boosted the place recognition. Radenović et al. [40] exploited the SfM result and automatically generated similar and non-similar image pairs, which is used to fine tune pre-trained CNNs for better global image features. Based on [40], Shen et al. [41] adjusted CNN by considering the local overlapping regions. Recently, Hou et al. [26] proposed a CNN fine-tuning method with multiple NetVLADs to aggregate feature maps of various channels and published an benchmarks *LOIP* that consists of both crowdsourced and photogrammetric images.

### C. Efficient optimization of bundle adjustment

Nowadays, bundle adjustment (BA) has become a mature technique for optimizing image poses and 3D point positions. However, as image number increases, a lot of works for solving BA in a fast and reliable way emerged. For example, preconditioned conjugate gradients were explored to solve BA in [42], Wu et al. [43] and Zheng et al. [44] further improved the efficiency for solving large-scale linear equation system by means of GPU. To cope with large-scale problem, distributed approaches that split a large BA problem into several overlapping small subset BA problems attract researchers' attentions [45-46]. [45] parallelly solved each subset BA and proposed global camera consensus constraint to merge all subsets, [46] employed 3D points as global consensus constraints and the corresponding covariance information was applied for better convergence behavior. MegBA [47] parallelly solved subsets via multiple GPUs, which provide a more time efficient solution.

All the above BA methods aims to efficiently optimize all unknows globally, which are inherently not feasible for incremental or sequential mode (it is not efficient to run global BA when each and every new image comes in, see section IV-C).

## III. ON-THE-FLY SFM

In this section, we introduce our on-the-fly SfM in more detail. First, we overview the general pipeline of our SfM that can perform online SfM while capturing image in arbitrary manner. Then, three key enrolled methodologies are explained: 1) Fast image retrieval based on learning-based global feature and vocabulary tree; 2) Correspondence refinement using least squares matching; 3) Efficient BA optimization via weighted hierarchical tree.

### A. Overview of on-the-fly SfM

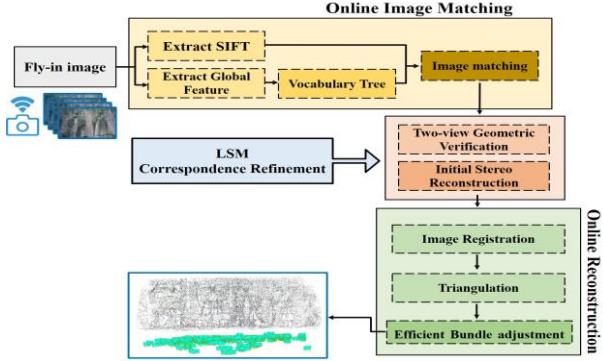

Figure 2. Workflow of the proposed on-the-fly SfM

Fig.2 illustrates the general workflow of our on-the-fly SfM, which constitutes five parts: image capturing and transmitter, online image matching, two-view geometry, LSM correspondence refinement, online reconstruction. Next, we explain each part.

**Image capturing and transmitter**. To achieve the goal of what you capture is what you get, in this work, a consumer digital camera is used to collect images, which is integrated with a wireless Wifi transmitter to transfer images for processing in real time (see section IV for more details). After receiving a new fly-in image, the other four parts start to work.

**Online image matching**. Fast identifying matchable images for new fly-in image is one of the most important procedures, as the first step for a new image is to find the relationship with already registered images, i.e., running image matching. In this paper, we applied the learning-based global feature [26] and its corresponding vocabulary tree to fast determine new image's matchable candidate images, among which correspondences are estimated.

**Two-view geometry**. Similar to [5], a multi-model two-view geometric verification method is applied. In general, fundamental matrix is estimated and two images are geometrically reliable if at least $N_f$ inlier matches exist, then the homography is computed with $N_h$ inliers. For calibrated case, essential matrix is estimated as well. And the final two-view geometric model is selected according to GRIC [48], and initial stereo reconstruction is selected as the verified image pair with most triangulated 3D points and the median triangulation angle being closed to 90 degrees (e.g., 60~120).

**LSM correspondence refinement**. Despite the employed robust estimator in two-view geometry and online reconstruction, a further improvement can be expected by refining the generated correspondences based on least squares matching.

**Online reconstruction**. This part mainly addresses on image pose and 3D point estimation, among which the image registration and triangulation are solved by EPnP [49] and RANSAC-based multi-view triangulation [5]. To approach online reconstruction, we solve the most time-consuming bundle adjustment by presenting hierarchical weighted local bundle adjustment which is based on the fact that newly fly-in image only affects its connected overlapping images to some degree (more details can be found in section III-D)

### B. Fast image retrieval based on learning-based global feature and vocabulary tree

In this part, a fast image retrieval pipeline integrated with learning-based global feature and vocabulary tree is employed to guarantee online image matching for on-the-fly SfM. Fig. 3 illustrates the key idea: 1. Pre-train models. CNN model is applied as global feature extractor [26,39,40], and a vocabulary tree is built using global features of all training images; 2. Image retrieval for fly-in image. Each new image's global feature is firstly extracted using selected CNN model, and input into built vocabulary tree to fast identify matchable images.

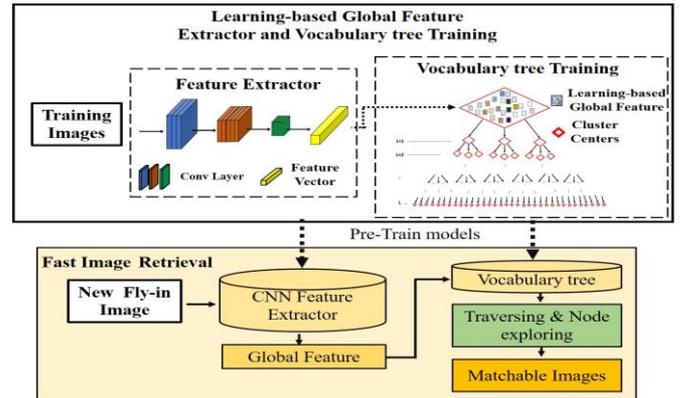

Figure 3. Fast image retrieval workflow based on learning-based global feature and vocabulary tree

1) *Learning-based Global Feature Extractor*. CNNs have been successfully applied in retrieving visually similar images as feature extractor [50]. In this work, to determine matchable image pairs that often have partial overlapping area, the fine-tuned CNN model of [26] is selected as our global feature extractor, as we find that [26] is tailored for seeking overlapping image pairs to speed up offline SfM and is supposed to be also feasible for our on-the-fly SfM. In particular, [26] yields a new training dataset (*LOIP*) with ground-truth matchable pairs, and a novel architecture composed of CNN and multiple NetVLADs are fine-tuned by region triplet loss. Note that their off-the-shelf model is accessible and employed.

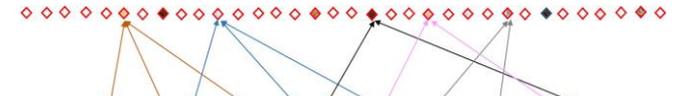

Figure 4. Toy example for fast image retrieval of new fly-in image. Similar images are clustered into the same node.

2) *Vocabulary tree training*. To the best of our knowledge, for global features, similar images are typically retrieved by comparing Euclidean distance of two images' feature vectors, which is yet not efficient for large scale problem. Motivated by BoW, it can be expected that a vocabulary tree for global feature is able to further improve retrieval time efficiency. Given the extracted global feature by [26], we can train a corresponding vocabulary tree via an unsupervised manner, i.e., the canonical K-means algorithm is hierarchically repeated to split the feature space until a certain depth is reached. To ensure the generality and even splitting of the feature space, *LOIP* containing various crowdsourced and photogrammetric images is used. As a consequence, a vocabulary tree with the information of each cluster center is generated for fast image retrieval.

3) *Fast Image Retrieval for new fly-in image*. Based on the pre-trained models of global feature extractor and vocabulary tree, matchable images of new fly-in image can be fast found. Instead of estimating Euclidean distance of all possible image pairs, only the cluster centers are required to be compared and the assumption is that similar images should fall into the same node as Fig. 4 shows. More specifically, as a new image flies in, its global feature is extracted and fed into the vocabulary tree, the already registered images that are matchable image candidates can be fast identified via traversing the nodes of vocabulary tree, i.e., similar images should always be in the same node.

### C. Correspondence Refinement using Least Squares Matching

Based on the original feature matching mechanism (e.g., SIFT), we present a correspondence refinement solution by integrating with the least squares matching (LSM), which is supposed to mitigate error accumulation. In general, as Fig. 5 shows, LSM is firstly applied to improve correspondences regarding 2D position and outliers, and to generate new observations for improving PnP estimation.

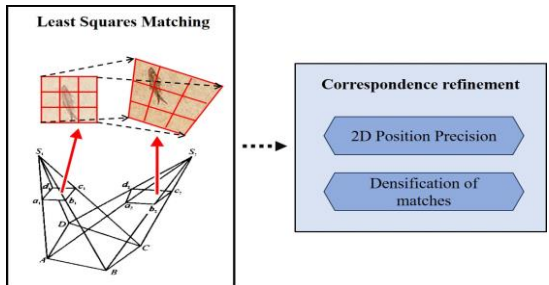

Figure 5. Least square matching refinement.

1) *Basic Principle of LSM*. The general idea of LSM is to optimize the 2D position of matches based on consistency of pixel grey values around corresponding local windows on two images [27]. Typically, radiometric and geometric inconsistency are explored in LSM, the first one often results from illumination, various photographic conditions and errors of digitization, etc., the second one is normally due to depth changes and image distortion, etc. The basic assumptions of LSM are: radiometric inconsistency between matched points is not complicated and can be approximated by linear transformation (see equation (1)) and the geometric inconsistency between two corresponding small local windows can be simply modelled by affine transformation (as Fig (5) left shows, see equation (2)). LSM is formulated by Equation (3) that combines Equation 1 and 2.

$$I_1(x_1, y_1) = h_0 + h_1 I_2(x_2', y_2') \quad (1)$$

$$\begin{cases} x_2' = a_0 + a_1 x_2 + a_2 y_2 \\ y_2' = b_0 + b_1 x_2 + b_2 y_2 \end{cases} \quad (2)$$

$$I_1(x_1, y_1) = h_0 + h_1 I_2(a_0 + a_1 x_2 + a_2 y_2, b_0 + b_1 x_2 + b_2 y_2) \quad (3)$$

where $I(.)$ indicates the grey value, $(x_1, y_1)$ and $(x_2, y_2)$ are the correspondence from original matching results. $a_{0\sim2}$ and $b_{0\sim2}$ are the unknown affine parameters, $h_{0,1}$ are the unknown linear parameters for radiometric constraint.

Equation (3) can be solved using least squares in an iterative way [27]. If the refinement converges successfully, the refined 2D position can be obtained from Equation (2), otherwise, the correspondence is deleted as outlier.

2) *2D position refinement and outlier detection*. According to the basic principle of LSM, given a pairwise correspondence, i.e., $(x_1, y_1)$ and $(x_2, y_2)$, we first try to solve equation (3) using least squares: if it converges, the corresponding 2D position will be refined; if it fails, the correspondence is detected as an outlier.

3) *Densifying matches*. For new fly-in image, one of the main goals is to compute the corresponding pose via EPnP. To ensure a robust and reliable pose estimation, new reliable extra 2D-3D matches are produced using LSM. For some 3D points that can be viewed by a specific image, but without corresponding 2D observations, LSM is run to generate these new 2D-3D matches. More specifically, initial pose is first estimated, 3D points are reprojected onto image for coarse 2D positions, LSM is then followed to optimize for more accurate 2D positions as densified matches. Finally, all the 2D-3D matches including both original and densified ones are employed for pose estimation.

### D. Hierarchical weighted local bundle adjustment for efficient optimization

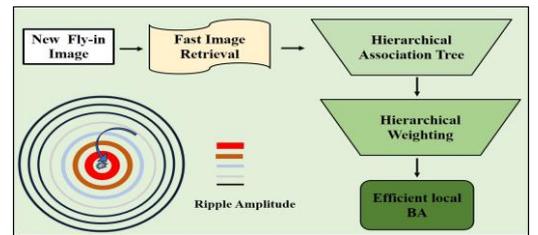

Figure 6. Hierarchical weighted local bundle adjustment.

To achieve real-time performance for our on-the-fly SfM, an efficient bundle adjustment is heavily required. Inspired by the natural phenomenon that the closer to center the ripple is, the larger the related amplitude is (as Fig.6 left corner shows), analogously, the uncertainty of new fly-in image makes higher influence on closely associated images than images that are farther. As Fig. 6 implies, this work presents a new efficient local bundle adjustment with hierarchical weights. Based on the image retrieval results (section *B*), a hierarchical association tree is built, which indicates the association relationship between new image and registered images. Then, hierarchical weight for every locally associated image is then estimated and used for robust bundle adjustment.

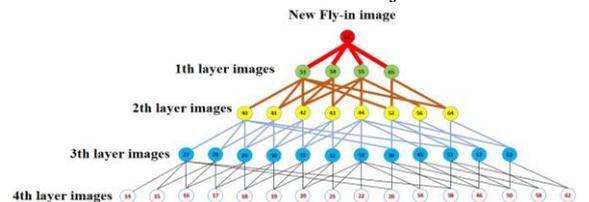

Figure 7. Example of hierarchical association tree building and weighting.

*1) Hierarchical association tree building and weighting.*
With the presented fast image retrieval method, for every fly-in image, it is efficient to figure out top-N similar images. As images on various ripples (or hierarchical layer) are inconsistently affected by new image, a *Hierarchical association tree* is built. The images in first ripple are composed of Top-N similar images for current new fly-in image, and the second ripple images are Top-N similar images of first ripple images, repeat until a pre-setting depth $L$ is reached. All the enrolled images in the hierarchical tree are denoted as $I^{hat}$. As Fig. 7 shows, a toy 4-layer hierarchical association tree is illustrated, in which every bottom layer images are the retrieved Top-N images of the upper layer and the first layer contribute highest effect on new image (indicated by thick red line). According to the ripple phenomenon, this work introduces a simple yet efficient hierarchical weighting solution for various ripple images, as shown in Equation (4):

$$p_i = \begin{cases} 1, & if\ i = * \\ (k)^{i-1}, & if\ i \neq L \\ \infty, & if\ i = L \end{cases} \quad (4)$$

where $i$ is the index of layer number, * is the current new fly-in image and $k$ is a constant value ($k>1$) denoting the basic inverse influence between new fly-in image and already registered images. The larger the $i$ is, the higher the corresponding $p_i$ is, which means images on farther ripples are much more stable and should have smaller updates.

*2) Local bundle Adjustment with hierarchical weights.*
Based on the local block consisting of $I^{hat}$ and weighting $p_i$, we establish a new efficient and robust local BA with hierarchical weights. Equation (5) denotes the original reduced normal equation with only camera parameters (see [6] for more details).

$$(J^T J + \lambda D^T D)\delta = -J^T f \quad (5)$$

To run bundle adjustment in a fast and robust way for new fly-in image, this study modifies Equation (5) as shown in Equation (6)

$$(J^T J + \lambda D^T D)P^{hat}\delta^{hat} = -J^T P^{hat} f \quad (6)$$

where, only the local block BA ($\delta^{hat}$) with images $I^{hat}$ are refined and reasonable weights $P^{hat}$ composed of corresponding $p_i$ is employed for robust optimization.

## IV. EXPERIMENTS

In this section, we report extensive experimental results on various datasets to demonstrate the capability of "*what you capture is what you get*" for our on-the-fly SfM.

### A. Implementation details

The learning-based global features are extracted by [26] and the vocabulary tree is trained with all images *LOIP* [26]. In Fig. 8, our online image transmission is integrated with CAMFI 3.0 wireless image transmission equipment, whose working area is around 50 meters and transmission speed can be up to 10 Mb/s. Typically, 3-5s are needed to receive one image since it is captured in our tests. All experiments are run on the machine with 16 CPU processors and RTX3080 GPU.

**Experimental datasets**. As fig. 8 shows, two self-collected datasets (*SX*-221 images, *YX*-349 images) are used to evaluate the on-the-fly performance of our SfM, which were taken in an arbitrary way and transferred online to our system. Three visual sequences (*fr1_desk*，*fr3_st_far*，*fr1_xyz*) from TUM RGB-D datasets [50] are simulatively employed as input.

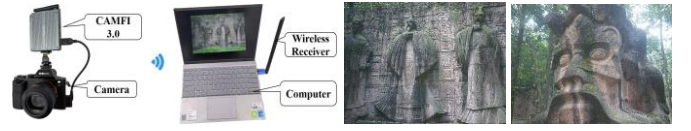

Figure 8. Online Image transmission (left - Hardware, middle-*SX*, right - *YX*).

**Running Parameters**. In this work, some free parameters are empirically set. For the online image matching, the vocabulary tree is with 5-layer depth and 5 sub-clusters for each node. Each new fly-in images selects Top-30 similar images for subsequent matching. The small local window in LSM is set as 15 ×15 pixels. For efficient BA, as each image in the ripple has top-N candidate images which might return a large BA block, only top-8 similar images are considered. The constant weighting parameter $k = 2$ in all experiments.

### B. Performance of fast image retrieval

To validate the real-time performance of our online image matching, based on *SX* and *fr3_st_far*, we investigate three different image matching strategies: exhaustive matching using Colmap with default setting (EM), exhaustive Euclidean comparison using learning-based global feature [26] (EE) and the proposed image retrieval (Ours) based on learning-based global feature and vocabulary tree.

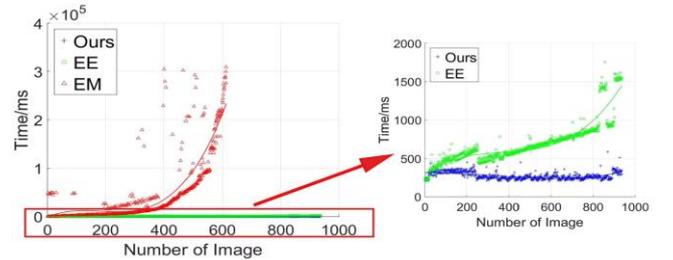

Figure 9. Time consuming of various methods on *fr3_st_far*.

Fig. 9 compares the time consumption of each strategy, in which the horizontal axis represents the number of current fly-in image and vertical axis is the time cost of retrieving current image with all the already registered images. It can be found that using global feature is significantly faster than the original local feature-based matching mechanism of Colmap, especially for large-scale dataset. In addition, when comparing with EE and Ours, the vocabulary tree can further improve time efficiency. The time cost of Ours is linear to the increasing number of images, while EE's time cost is quadratic.

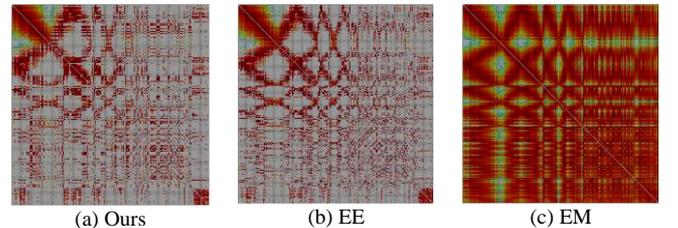

(a) Ours      (b) EE      (c) EM

Figure 10. Overlapping graph of *SX*. Vertical and horizontal axis are image ID. The darker red the pixel is, the higher possibility the corresponding image pair overlaps with each other.

Fig. 10 qualitatively shows the matching results that both Ours and EE can identify the basic skeleton of EM, which means the most similar images determined by EM are successfully found by Ours and EE.

### C. Performance of efficient local bundle adjustment

To demonstrate the efficacy of the presented local bundle adjustment, different bundle adjustment solutions are compared: first, a global bundle adjustment that enrolls all images is performed (Glo.); second, a combined solution integrated with local and global bundle adjustment (Com.), this is actually successfully applied in Colmap [5]; third, local bundle adjustment with hierarchical weights with (Ours). Based on *fr3_st_far*, these three bundle adjustment solutions are tested for BA optimization when a new image comes into the block.

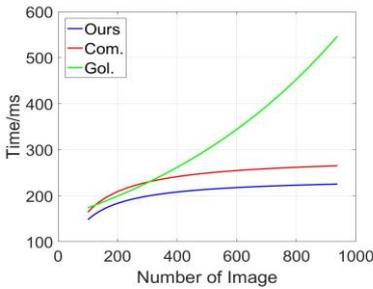

Figure 11. Cost time on *fr3_st_far* with various bundle adjustment methods.

Fig. 11 shows the time cost for different BA methods, which records the optimization time for each new fly-in image. It can be found that, as the image number grows, the consuming time increases dramatically for the Global method, the cost time of Ours increases the slowest and tends to be stable after adding some images. This can be explained by the fact that, as more images involve, more time is needed to refine more unknown parameters. The whole block is considered for Global method, whereas, Ours only solves a local bundle adjustment for images in the built hierarchical association tree. Tab. I lists quantitative results, i.e., averaging mean reprojection error of each BA (AMRE), mean reprojection error of final BA (MFRE) and mean track length (MLT), these results are nearly similar and in the same magnitude. Therefore, the presented BA is fast yet robust solution, and is feasible to our on-the-fly SfM.

TABLE I. QUANTITATIVE RESULTS OF VARIOUS BA METHODS

| Dataset | AMRE | | | MFRE | | | MTL | | |
|---|---|---|---|---|---|---|---|---|---|
| | Ours | Com. | Glo. | Ours | Com. | Glo. | Ours | Com. | Glo. |
| fr3_st_far | 0.59 | 0.53 | 0.48 | 0.33 | 0.33 | 0.35 | 35.53 | 34.66 | 32.02 |

### D. On-the-fly performance of our SfM

TABLE II. COST TIME OF EACH CORE STAGE IN OURS SFM (MS)

| Dataset | SX | YX | fr1_desk | fr1_xyz | fr3_st_far |
|---|---|---|---|---|---|
| NoI | 221 | 349 | 613 | 798 | 938 |
| FE | 617 | 625 | 157 | 155 | 172 |
| OIM | 1282 | 1391 | 872 | 911 | 1247 |
| GV | 1285 | 951 | 168 | 187 | 359 |
| IR | 91 | 72 | 41 | 56 | 72 |
| Tri. | 158 | 171 | 116 | 29 | 60 |
| BA | 190 | 131 | 74 | 184 | 198 |
| Total | 3623 | 3341 | 1428 | 1522 | 2108 |
| IT | 4200 | 4400 | 3500 | 3500 | 3500[a] |

a. It is simulated that 3.5s is needed for each video frame transmission.

Tab. II presents the average processing time for all images of each dataset, in particular, several key procedures are reported: image transmission (IT), feature extraction (FE), online image matching (OIM), two-view geometric verification (GV), Image registration (IR), Triangulation (Tri.) and bundle adjustment (BA). We can find that OIM and GV take the most time, and all the others are quite fast. It is worth noting that, basically, for our on-the-fly SfM, current fly-in image can be solved before next image is received.

### E. Comparison with other state-of-the-art SfM

In addition, to further explore how far is our SfM to the state-of-the-art SfM, we make comparative investigation involving two popular SfM systems, namely, Colmap [5] and OpenMVG [30]. Due to that both Colmap and OpenMVG are only with offline mode, time efficiency is not discussed here. Our SfM results of *SX* and *YX* are visualized in Fig. 12.

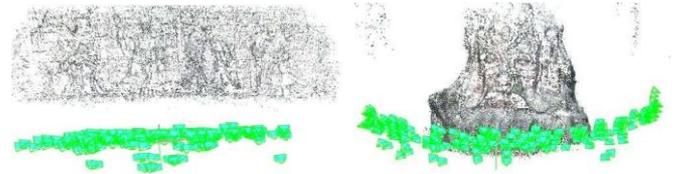

Figure 12. Visualization of our SfM results on *SX* and *YX*.

TABLE III. COMPARISON BETWEEN ON-THE-FLY SFM AND COLMAP

| Dataset | MRE | | MTL | | Rotation Discrepancy (in degrees) |
|---|---|---|---|---|---|
| | Ours | Colmap | Ours | Colmap | |
| SX | 0.84 | 0.90 | 11.43 | 13.42 | 0.05 |
| YX | 0.65 | 0.80 | 9.59 | 11.55 | 0.25 |
| fr1_desk | 0.60 | 0.72 | 12.65 | 15.45 | 0.40 |
| fr1_xyz | 0.63 | 0.71 | 28.68 | 45.17 | 0.79 |
| fr3_st_far | 0.33 | 0.61 | 35.53 | 62.60 | 0.07 |

TABLE IV. COMPARISON BETWEEN ON-THE-FLY SFM AND OPENMVG

| Dataset | MRE | | MTL | | Rotation Discrepancy (in degree) |
|---|---|---|---|---|---|
| | Ours | OPENMVG | Ours | OPENMVG | |
| SX | 0.84 | Fail | 11.43 | Fail | Fail |
| YX | 0.65 | 0.67 | 9.59 | 6 | 0.24 |
| fr1_desk | 0.60 | 0.92 | 12.65 | 8 | 1.16 |
| fr1_xyz | 0.63 | 0.86 | 28.68 | 12 | 0.72 |
| fr3_st_far | 0.33 | 0.63 | 35.53 | 16 | 0.27 |

In Tab. III and IV, three criteria are studied including mean reprojection error, mean track length and rotation discrepancy (taking Colmap and OpenMVG as reference). In general, the MRE values from our SfM, Colmap and OpenMVG are less than 1 pixel, which typically demotes a converge behavior in BA, in most cases, we obtain better MRE which might be resulted from the refined correspondences from our least squares matching. The final MTL varies a lot, this is due to that they used various image matching packages and outlier detection strategies in BA. The very small rotation discrepancy on rotation results shows that our SfM is capable to yield considerable camera poses as these popular SfM packages are.

## V. CONCLUSION

In this work, we present a novel on-the-fly SfM: running online SfM while image capturing, which achieves the goal of *what you capture is what you get*. Three technical improvements of learning-based online image matching, correspondence refinement using least squares and efficient local bundle adjustment using hierarchical weights are employed to guarantee a fast yet robust online SfM. Extensive results of various datasets demonstrate the real-time performance and robustness of our on-the-fly SfM.

REFERENCES

[1] W. Liu et al., "Ground Camera Images and UAV 3D Model Registration for Outdoor Augmented Reality," in *IEEE Conference on Virtual Reality and 3D User Interfaces (VR)*, 2019, pp. 1050-1051.
[2] P. Sarlin, A. Unagar, M. Larsson, H. Germain, C. Toft, V. Larsson, M. Pollefeys, V. Lepetit, L. Hammarstrand, F. Kahl, and T. Sattler, "Back to the feature: Learning robust camera localization from pixels to pose," in *IEEE Conference on Computer Vision and Pattern Recognition*, 2021, pp. 3247–3257.
[3] P. Sarlin, C. Cadena, R. Siegwart, and M. Dymczyk, "From coarse to fine: Robust hierarchical localization at large scale," in *IEEE Conference on Computer Vision and Pattern Recognition (CVPR)*, 2019, pp.12716–12725.
[4] E. Brachmann, M. Humenberger, C. Rother, and T. Sattler, "On the limits of pseudo ground truth in visual camera re-localisation," in *IEEE/CVF International Conference on Computer Vision (ICCV)*, 2021, pp. 6218–6228.
[5] J. L. Schönberger and J. -M. Frahm, "Structure-from-Motion Revisited," in *IEEE Conference on Computer Vision and Pattern Recognition (CVPR)*, 2016, pp. 4104-4113.
[6] C. Wu, "Towards Linear-Time Incremental Structure from Motion," in *IEEE Conference on 3DTV*, 2013, pp.127-134.
[7] S. Agarwal, N. Snavely, I. Simon, S. Seitz, R. Szeliski, "Building Rome in a day," in *IEEE Conference on Computer Vison*, pp. 72-79.
[8] J.M. Frahm, P. Fitegeorgel, D. Gallup, T. Johnson, R. Raguram, C. Wu, et al., "Building Rome on a cloudless day," in *European Conference on Computer Vison*, pp. 368-381.
[9] X. Wang, F. Rottensteiner, C. Heipke, "Robust image orientation based on relative rotations and tie points," in *ISPRS Ann. Photogram., Rem. Sens. Spatial Inf. Sci. IV-2,* 2018, pp.295–302.
[10] R. Gherardi, M. Farenzena, A. Fusiello, "Improving the efficiency of hierarchical structure-and-motion," in *IEEE Conference on Computer Vision and Pattern Recognition (CVPR)*, 2010, pp. 1594-1600.
[11] R. Toldo, R. Gherardi, M. Farenzena, A. Fusiello, "Hierarchical Structure-and-motion recovery from uncalibrated images," in *Computer Vision & Image Understanding*, 2015, pp.127-143.
[12] M. Farenzena, A. Fusiello and R. Gherardi R, "Structureand-motion pipeline on a hierarchical cluster tree," in *IEEE International Conf. on Computer Vision (ICCV) Workshop*, 2009, pp.1489-1496.
[13] M. Havlena, A. Torii, J. Knopp, T. Pajdla, "Randomized structure from motion based on atomic 3D models from camera triplets," in *IEEE Conf. on Computer Vision and Pattern Recognition (CVPR)*,2009, pp.2874-2881
[14] N. Jiang, Z. Cui, P. Tan, "A global linear method for camera pose registration," in *IEEE International Conf. on Computer Vision (ICCV)*, 2013, pp.481–488
[15] Z. Cui, P. Tan, "Global Structure from Motion by Similarity Averaging," in *IEEE International Conference on Computer Vision (ICCV)*,2015, pp. 864-872.
[16] K. Wilson, N. Snavely, "Robust Global Translations with 1DSfM", *European Conference on Computer Vsion (ECCV)*, 2014, pp. 61-75.
[17] Y. Kasten, A. Geifman, M. Galun, R. Basri, "Algebraic Characterization of Essential Matrices and Their Averaging in Multiview Settings," in *IEEE International Conf. on Computer Vision (ICCV)*, 2019.
[18] B. B. Zhuang, L.F. Cheong, G.H. Lee, "Baseline Desensitizing in Translation Averaging," in *IEEE Conference on Computer Vision and Pattern Recognition (CVPR),* 2018.
[19] F. Arrigoni, A. Fusiello, B. Rossi, "Camera motion from group synchronization," in *IEEE International conf. on 3D Vision*, 2016, pp.546-555.
[20] M. Arie-Nachimson, S.Z. Kovalsky, I. Kemelmacher-Shlizerman, A. Singer and R. Basri, "Global motion estimation from point matches," *In Proceedings of the IEEE Conf. on 3D Vision*,2012, pp. 81-88.
[21] Raúl Mur-Artal, J. M. M. Montiel and Juan D. Tardós, "ORB-SLAM: A Versatile and Accurate Monocular SLAM System," in *IEEE Transactions on Robotics*, 2015, pp. 1147-1163.
[22] D. Kiss-Illés, C. Barrado, E. Salamí, "GPS-SLAM: an augmentation of the ORB-SLAM algorithm," in *Sensors*, 2019, pp. 1-22.
[23] T. Qin, J. Pan, S. Cap, S. Shen, "A General Optimization-based Framework for Local Odometry Estimation with Multiple Sensors," in *arXiv:1901.03642*, 2019.
[24] R. Mur-Artal and J. D. Tardós, "ORB-SLAM2: An open-source SLAM system for monocular, stereo, and RGB-D cameras," *IEEE Trans. Robot.*, 2017, *vol. 33, no. 5*, pp. 1255–1262.
[25] C. Campos, R. Elvira, J. J. G. Rodríguez, J. M. M. Montiel and J. D. Tardós, "ORB-SLAM3: An Accurate Open-Source Library for Visual, Visual–Inertial, and Multimap SLAM," in *IEEE Transactions on Robotics*, 2021, pp. 1874-1890.
[26] Q. Hou, R. Xia, J. Zhang, et al., "Learning visual overlapping image pairs for SfM via CNN fine-tuning with photogrammetric geometry information," in *International Journal of Applied Earth Observations and Geoinformation*, 2023, 103162.
[27] Y. Yue, X. Wang and Z. Zhan, "Single-Point Least Square Matching Embedded Method for Improving Visual SLAM," in *IEEE Sensors Journal*, 2023, pp. 16176-16188.
[28] D. Lowe, "Distinctive image features from scale invariant keypoints," in *International Journal of Computer Vison (IJCV), 2004, vol .60, no. 2*, pp. 91-110.
[29] VisualSFM: http://ccwu.me/vsfm.
[30] OpenMVG: https://github.com/openMVG/openMVG.
[31] Theia: http://theia-sfm.org/.
[32] S. Song and M. Chandraker, "Robust Scale Estimation in Real-Time Monocular SFM for Autonomous Driving," in *IEEE Conference on Computer Vision and Pattern Recognition*, 2014, pp. 1566-1573.
[33] Y. Zhao, et al., "RTSfM: Real-Time Structure From Motion for Mosaicing and DSM Mapping of Sequential Aerial Images With Low Overlap," in *IEEE Transactions on Geoscience and Remote Sensing*, 2022, pp. 1-15.
[34] D. Nister, H. Stewenius, "Scalable recognition with a vocabulary tree," in *IEEE Conference on Computer Vision and Pattern Recognition (CVPR)*, 2006, pp. 2161-2168.
[35] P. Geneva, K. Eckenhoff, W. Lee, Y. Yang, and G. Huang, "OpenVINS: A research platform for visual-inertial estimation," in *IEEE Int. Conf. Robot. Autom. (ICRA)*, 2020, pp. 4666–4672.
[36] N. Snavely, S.M. Seitz, R Szeliski, "Photo Tourism: Exploring Photo Collection in 3D," in *ACM Transactions on Graphics*, 2006, pp. 835-846.
[37] M. Havlena, K. Schindler, "VocMatch: efficient multiview correspondence for structure from motion," in *European Conference on Computer vision*, 2014.
[38] X. Wang, F. Rottensteiner, C. Heipke, "Structure from motion for ordered and unordered image sets based on random k-d forests and global pose estimation" in *ISPRS Journal of Photogrammerty and Remote Sensing*, 2019, pp. 19-41.
[39] R. Arandjelović, P. Gronat, A. Torii, T. Pajdla and J. Sivic, "NetVLAD: CNN Architecture for Weakly Supervised Place Recognition," in *Conference on Computer Vision and Pattern Recognition (CVPR),* 2016, pp. 5297-5307.
[40] F. Radenović, G. Tolias and O. Chum, "Fine-Tuning CNN Image Retrieval with No Human Annotation," in *IEEE Transactions on Pattern Analysis and Machine Intelligence*, 2019, pp. 1655-1668.
[41] T.W. Shen, Z.X. Luo, L. Zhou, R.Z., Zhang, S.Y. Zhu, L. Quan, "Matchable Image Retrieval by Learning from Surface Reconstruction," in *Asian Conference on Computer Vision*, 2018, pp. 415-431.
[42] S. Agarwal, N. Snavely, S.M. Seitz, R. Szeliski, "Bundle Adjustment in the Large," in *European Conference on Computer Vision (ECCV)*, 2010, pp. 29-42.
[43] C. Wu, S. Agarwal, B. Curless and S. M. Seitz, "Multicore bundle adjustment," in *IEEE Conference on Computer Vision and Pattern Recognition (CVPR)*, 2011, pp. 3057-3064.
[44] M.T. Zheng, S.P. Zhou, X.D. Xiong, J.F. Zhu, "A new GPU bundle adjustment method for large-scale data," in *Photogrammetric Engineering & Remote Sensing*, 2017, pp.23-31.
[45] R.Z. Zhang, S.Y. Zhu, T. Fang, L. Quan, "Distributed Very Large Scale Bundle Adjustment by Global Camera Consensus," in *IEEE International Conference on Computer Vision*, 2017, pp. 29-38.
[46] H. Mayer, "RPBA-Robust parallel bundle adjustment based on covariance information," in *arXiv preprint arXiv:* 1910.08138, 2019.
[47] J. Ren, W.T. Liang, R. Yan, M. Luo, S.W. Liu, X.Liu, "MegBA: A GPU-Based Distributed Library for Large-Scale Bundle Adjustment," in *European Conference on Computer Vision*, 2022.

[48] P.H. Torr, "An assessment of information criteria for motion model selection," *in IEEE Conference on Computer Vision and Pattern Recognition (CVPR)*, 1997.

[49] V. Lepetit, F. Moreno-Noguer, P. Fua, "EPnP: An accurate O(n) solution to the PnP problem," *International Journal of Computer Vision*, 2009, *vol.81*, pp. 155-166.

[50] J. Sturm, N. Engelhard, F. Endres, W. Burgard, and D. Cremers, "A benchmark for the evaluation of RGB-D SLAM systems," *in IEEE International Conference Intelligent. Robots System*, 2012, pp. 573–580.